\documentclass[journal ]{IEEEtran}
\usepackage{graphicx,cite,epsfig,amssymb,amsmath,subfigure,color}

\begin{document}

\title{Testing Large Language Models on Driving Theory Knowledge and Skills for Connected Autonomous Vehicles}
\author{Zuoyin~Tang, Jianhua~He, Dashuai~Pei, Kezhong~Liu and Tao~Gao
\thanks{Zuoyin~Tang (email: {\tt z.tang1@aston.ac.uk}) is with School of Informatics and Digital Engineering, Aston University, UK. 
Jianhua~He (email: {\tt j.he@essex.ac.uk}, Corresponding Author) is with the School of Computer Science and Electronic Engineering, Essex University, UK. 
Dashuai~Pei (email: {\tt Pei.Dashuai@whut.edu.cn}) is with School of Computer Science and Electronic Engineering, University of Essex, UK, 
and School of Navigation, Wuhan University of Technology, China.
Kezhong~Liu (email: {\tt kzliu@whut.edu.cn}) are with School of Navigation, Wuhan University of Technology, China.
Tao~Gao (email: {\tt gtnwpu@126.com}) is with School of Information Engineering, Chang'An University, China.}}

\maketitle

\begin{abstract}
Handling long tail corner cases is a major challenge faced by autonomous vehicles (AVs).
While large language models (LLMs) hold great potentials to handle the corner cases with excellent generalization and explanation capabilities
and received increasing research interest on application to autonomous driving, there are still technical barriers to be tackled, 
such as strict model performance and huge computing resource requirements of LLMs, which are difficult to be met locally at AVs. 
In this paper, we investigate a new approach of applying remote or edge LLMs to support autonomous driving.
With this approach connected autonomous vehicles (CAVs) send driving assistance requests to the LLMs. 
LLMs deployed at the edge of the networks or remote clouds process the requests and generate driving assistance instructions for the CAVs. 
A key issue for such LLM assisted driving system is the assessment of LLMs on their understanding of driving theory and skills,
ensuring they are qualified to undertake safety critical driving assistance tasks for CAVs.
As there is no published work on assessing LLM of driving theory and skills, 
we design and run driving theory tests for several proprietary LLM models (OpenAI GPT models, Baidu Ernie and Ali QWen) 
and open-source LLM models (Tsinghua MiniCPM-2B and MiniCPM-Llama3-V2.5) with more than 500 multiple-choices theory test questions.
These questions are close to the official UK driving theory test ones.
Model accuracy, cost and processing latency are measured from the experiments.
Experiment results show that while model GPT-4 passes the test with improved domain knowledge
and Ernie has an accuracy of 85\% (just below the 86\% passing threshold),
other LLM models including GPT-3.5 fail the test.
For the test questions with images, the multimodal model GPT4-o has an excellent accuracy result of 96\%,
and the MiniCPM-Llama3-V2.5 achieves an accuracy of 76\%. 
While GPT-4 holds stronger potential for CAV driving assistance applications, 
the cost of using model GPT4 is much higher, almost 50 times of that of using GPT3.5.
The results can help make decision on the use of the existing LLMs for CAV applications and balancing on the model performance and cost.
\end{abstract}


\section{INTRODUCTION}\label{sec:introduction}

Road safety is a global challenge with at least one million people dying on the roads globally every year and millions more seriously injured \cite{Who23}.  
Autonomous vehicles (AVs) as a part of the safe vehicles dimension hold great potentials on improving driving safety.  
In the last several years there were significant advances on autonomous driving technologies. 
Field trials of self-driving vehicles have been demonstrated and robotaxis services have been provided in some cities. 
However, it still faces many technical challenges, such as the sensor limitations, lack of other vehicles' driving intention, 
incompetent on handling complex driving environments and corner cases \cite{He20}. 
For example, most AV sensors such as cameras and Lidars are limited to line-of-sight sensing 
and their performance can be largely affected by weather and light conditions. 
Furthermore, the deep learning models trained with large datasets for perception of driving environment and path planning may not see some unusual scenes and can't handle them due to low generalization capabilities.

\begin{figure*}[htb]
\centering		
\includegraphics[width= .654\textwidth]{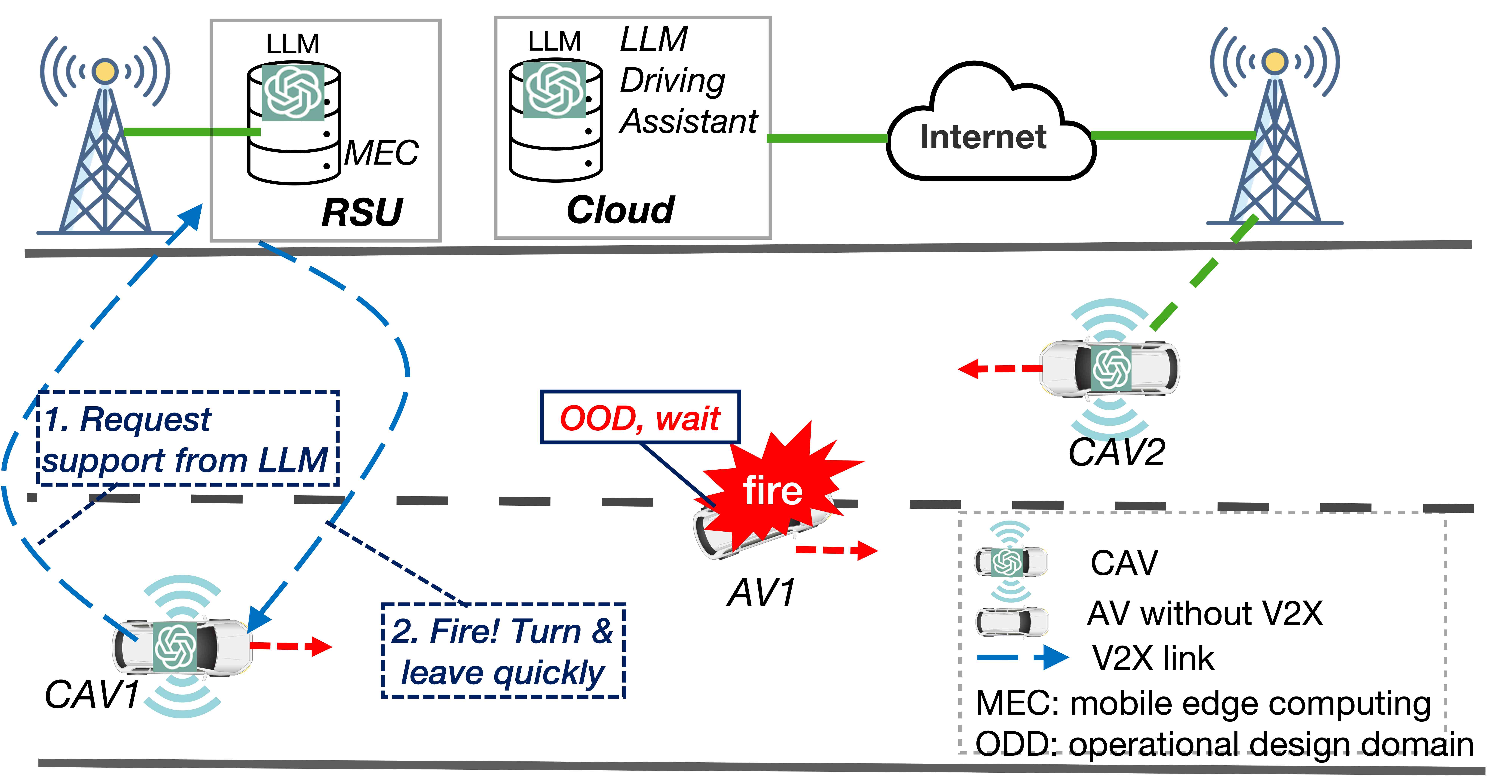}
\caption{System diagram of LLM driving assistance for CAVs. CAVs may have local LLMs or get assistance from remote LLMs.}
\label{fig:drivellm_system}	 		
\end{figure*}

\subsection{Existing works on LLM empowed autonomous driving}
To address the big challenge of handling corner cases, applying large language models (LLMs) to support autonomous driving 
is receiving increasing research interest by exploiting the excellent generalization abilities of LLMs.

\subsubsection{LLM for annotation}
D. Wu et al applied LLM for annotation of datasets for applications of capturing objects from driving scenes following flexible human commands \cite{Wu23}, similar to that used by human driving.

\subsubsection{LLM for end to end driving}
J. Mao et al proposed an approach that transforms the OpenAI GPT-3.5 model into a reliable motion planner for autonomous vehicles \cite{Mao23}.
Driving environment perception and prediction and the ego vehicle states are converted to language prompt.
The LLM model was fine-tuned to produce a planned trajectory alongside its decision-making process in natural language.

An interpretable end-to-end autonomous driving system utilizing LLMs was proposed \cite{Xu23},
which can interpret vehicle actions and provide  reasoning and  answer diverse questions from human users for enhanced interaction. Additionally, the proposed system DriveGPT4 can predict vehicle low-level control signals in an end-to-end fashion. 

H. Sha et al employed LLMs as a decision-making component for complex autonomous driving scenarios. 
They developed algorithms for translating LLM decisions into actionable driving commands
and LLM decisions are integrated with low-level controllers \cite{Sha23}.

\subsubsection{LLM for autonomous driving with knowledge and memory}
L. Wen et al proposed a knowledge-driven approach for autonomous driving with LLM, which can instil knowledge to the system
and integrate a memory component to accumulate experience. The proposed approach demonstrated an improved generalization
ability over reinforcement learning-based methods \cite{Wen23}\cite{Fu24}.

\subsubsection{LLM for specific driving tasks}
C. Cui et al proposed a new framework that utilizes LLM linguistic and contextual understanding abilities
to support driving tasks (such as lane merging) and improve safety,
with personalized driving experiences through ongoing verbal feedback \cite{Cui24}. 
It aims to achieve user-centric, transparent, and adaptive autonomous driving.

More existing works on LLM for autonomous driving can be found from the survey papers \cite{Zhou23}\cite{Cui24WACV}.

\subsection{Research motivation}
While there have been many interesting works attempting to apply LLMs for autonomous driving,
these works mainly focus on using local LLMs and there are still several major technical challenges.
Generally, LLMs need more parameters to have a higher capacity to capture complex patterns 
in the training data and potentially have better performance. 
These models will have substantial computation and memory requirements due to their large parameter size,
which are difficult to be satisfied by embedded driving systems at AVs.
Furthermore, autonomous driving is safety critical with much higher safety expectation than human driving.
While human drivers are required to pass both theory and practical road tests before they are qualified for driving on the public roads,
there is no strict assessment of the LLMs for autonomous driving.
The reported safety performance of the LLM empowered autonomous driving is much lower than that of human drivers \cite{Wen23}.
Therefore, there are important questions which are not addressed:
\begin{itemize}
\item How to efficiently and safe use LLM for CAVs?
\item How much driving knowledge and skills do LLMs understand?
\item How well is LLMs qualified and how much can we trust them for driving and/or assistance according to official driving theory test?
\end{itemize}
In view of the research potentials and challenges of LLMs for CAV driving or assistance,
we are motivated to propose an approach of applying remote or edge LLMs to support autonomous driving.
With this approach connected autonomous vehicles (CAVs) send requests of driving assistance to the LLMs. 
The LLMs are deployed at the edge of the networks or remote clouds. 
They process the request and generate driving assistance for the CAVs. 
While connected autonomous vehicles (CAVs) has been considered as a key to address the challenges faced 
by autonomous driving and is widely studied \cite{Noo22},
to the best of our knowledge, application of LLMs for CAVs has rarely been studied. 

In addition, we aim to assess the LLMs on their knowledge of driving theory and skills in a way to human driver test,
before they can be used to undertake safety critical driving assistance services for CAVs.
Compared to the practical driving skill test on roads or through simulations to test the LLMs' capabilities of perception and controlling vehicles,
we believe driving theory test of LLMs are also very important and relatively easy and controllable.
While LLMs have been tested and demonstrated impressive performances over different fields such as laws, education and economics, 
driving theory test of LLMs has been rarely reported.
In this paper we design and run driving theory test  for several proprietary LLM models (OpenAI GPT models, Baidu Ernie and Ali QWen) 
\footnote{https://openai.com/}
and open-source LLM models (Tsinghua MiniCPM-2B and MiniCPM-Llama3-V2.5).
These questions are close to those used in the official UK driving theory tests.
Model accuracy, cost and processing latency are measured from the experiments.

Experiment results show that while model GPT-4 passes the test with improved domain knowledge
and Ernie has an accuracy of 85\% just below passing threshold,
other LLM models including GPT-3.5 fail the test.
For the test questions with images, the multimodal model GPT-4o has an excellent accuracy result of 96\%,
and the MiniCPM-Llama3-V2.5 achieves an accuracy of 76\%. 
It is also noted that while GPT-4 holds stronger potential for CAV driving assistance applications, 
the cost of using model GPT-4 is much higher, almost 20 times that of using GPT-3.5.
The research reported in this paper can help make decisions on using the LLMs for CAVs with a balance on model performance and cost.
The theory test framework is also applicable to driving assistance for AVs with locally deployed LLMs.

\section{System Framework of LLM Assisted Driving for CAVs}

In contrast to the most existing works on applying local LLMs deployed in vehicles for real time autonomous driving,
we investigate a new approach of exploiting LLMs deployed in remoted clouds or network edges to assist CAV driving.
A system framework diagram is presented in Fig.\ref{fig:drivellm_system}.
In the LLM assisted driving system, CAVs communicate with LLMs via the vehicle to everything (V2X) technologies.
For example, they can communicate with LLMs deployed at the roadside unit (RSU) through direct vehicle to vehicle (V2V) communication,
or via vechile to infrastructure (V2I) links to cellular base stations and then connect to RSUs.
Alternatively, CAVs may communicate with LLMs deployed at remote clouds via V2I links to cellular base stations and Internet. 

The LLMs may be the general ones provided from the leading LLM companies, such as OpenAI, Anthrotic, Google and Meta.
These models include both single modal ones (such as text based ChatGPT model) 
and multi-modal ones which can process text, image and/or audio inputs (such as GPT-4o, Claude3 and Gemni Pro).
While these LLMs are not trained specifically for CAV driving tasks, 
their rich knowledge and strong generalization abilities obtained from training over huge corpus may perform well for the tasks.
Furthermore, customized LLMs may be trained for the CAV driving assistance tasks,
which can have better understanding of the domain knowledge and applications, and provide reliable and robust assistance services to the CAVs.

There can be a wide range of autonomous driving assistance services that are provided by the LLMs on top of CAV applications,
which include remote driving with or without road perception at the LLMs, emergency response, occasional driving assistance at complex road or driving scenarios.
With CAV technologies, they can share their status, sensing and driving intents to cooperate on environment perception and driving. Third Generation Partnership Project (3GPP) has specified advanced driving uses with enhanced V2X communications \cite{3GP22}. These uses include cooperative sensing, vehicle platooning, remote driving and cooperative driving \cite{3GP22}. The LLMs can be used to support all these advanced driving uses and other driving assistance applications.

One driving assistance example with LLM is shown in Fig.~\ref{fig:drivellm_system}.
One AV (shown as AV1) caught fire and stopped in the middle of road out of the operational design domain (ODD).
Two CAVs (CAV1 and CAV2) went by and didn't know how to deal with the road accident.
CAVs sent a request to a LLM located at the nearby RSU for driving assistance.
The LLM replied to CAV1 with an instruction of turninging and leaving quickly. 
CAV1 followed the LLM instruction to avoid damage and further accidents.

While the LLM assisted driving holds great potential for safe and efficient CAV driving,
there are some major challenges to be tackled.
Firstly, the CAV driving tasks are safety and latency critical. 
The remote deployment of LLM may incur excessive communication and computation time,
which can have large impact on the driving assistance services, such as remote driving and emergency response services.
The feasibility and performance guarantee issues will need to be studied further for the application LLM to CAV driving.
Secondly, LLMs are known to have hallucinations which refer to outputs generated by LLMs that are convincingly wrong or misleading.
This can present significant risks and adverse impact on CAV driving.
Furthermore, the human drivers are required to qualify for driving with a driving licence obtained by passing the driving test.
This should be applied to the AI drivers or LLM driving assistants,
which will be studied in the next section. 

\section{Research Methodology}

\subsection{UK Driving Theory Test}

In this section we design theoretic driving test for LLMs to examine how well they understand the highway code and traffic signs.
For human drivers, they are required to pass theory test to prove they are qualified with right knowledge, understanding and attitude to be a safe and responsible driver \cite{DVSA24}. In the UK, the driving theory test include two parts, 1) a multiple choice test to assess the knowledge of driving theory and highway code (such as the road rules and driving practice), and 2) a hazard perception test to assess the hazard recognition skills. In this paper we will mainly focus on the UK multiple choice test to assess driving theory knowledge. The hazard perception test is left for our future work.

In the UK driving multiple choice test there are 50 multiple choice questions, each with 4 choices and one or more correct answers. 
The test questions cover a variety of topics relating to road safety and environment. 57 minutes are given to human learn drivers for this multiple choice test. To pass the multiple choice part of the theory test, at least 43 out of the 50 test questions must be answered correctly for human learner car drivers. This pass criteria of 86\% accuracy will also be applied to the LLMs to be used for driving assistance.

\subsection{Test Datasets}
In a book published by the Driver \& Vehicle Standard Agency (DVSA) \cite{DVSA24}, 
all the questions that may be present in the multiple choice test are included. 
These questions are ideal for testing the LLMs.
However, as we don't have access to the electronic version of these test questions,
we use alternative test questions publicly available online from the DriverInstructor website
\footnote{https://www.drivinginstructorwebsites.co.uk/uk-driving-theory-test-practice-questions-and-answers, accessed on 1 June 2024.},
which are very close to the official DVSA theory questions.
There are about 640 test questions in the DriverInstructor website after excluding the repeated questions.
Among these questions, there are about 50 questions with images such as traffic scenes and traffic signs.

As some of the LLMs to be tested can't process image input data,
we create two datasets from the DriverInstructor test questions, which include test questions with and without images, respectively.
In the dataset with questions without images (denoted as DS-Text), there are 544 test questions;
in the dataset with  questions with images (denoted as DS-Image), there are 53 questions.

\subsection{LLMs Used in Theory Test}
There are several powerful LLMs from leading companies such as Anthropic, Google, OpenAI, Ali and Baidu.
Several proprietary LLMs from OpenAI, ALi and Baidu are chosen for driving theory test in this paper, which are among the best performing ones.

\subsubsection{OpenAI Models}
The three selected OpenAI models (GPT-3.5 Turbo, GPT-4 and GPT-4o) have different capabilities and price points \footnote{https://openai.com/api/pricing/, last accessed on 10 June 2024}. 
GPT-3.5 Turbo (called GPT-3.5 for simplicity in the remaining of the paper) is a fast and inexpensive model for simpler tasks.
It supports 16K context window and is optimized for dialog.
GPT-4 was built with broad general knowledge and domain expertise, which shows much stronger performance in the driving theory test.
But GPT-4 model is much more expensive with a input price 60 times of that for GPT-3.5.
GPT-4o is OpenAI's most advanced multimodal model, 
which is faster and cheaper than GPT-4, and has stronger vision capabilities. The model has 128K context.

The OpenAI API is used to call the LLMs with input of test questions and obtain the model prediction output.
Table~\ref{tab:llm-price} show the price points of the used OpenAI LLMs in the units of 1M tokens, on 1 June 2024.
\begin{table}[ht]
    \centering
    \caption{Price points of OpenAI LLMs for input and output (units of 1M tokens) and image input (resolution of 150px by 150 px).}
    \begin{tabular}{llll}
    \hline
        Model name &  Input price & Output price  & Image  \\ \hline
        GPT-3.5  & \$0.50  & \$1.50  & N/A\\
        GPT-4 &  \$30.00  & \$120.00 & N/A \\
        GPT-4o  & \$5.0  & \$15.00  & \$0.001275 per image  \\  
   \hline
    \end{tabular}
    \centering
    \label{tab:llm-price}
\end{table}

\subsubsection{Alibaba and Baidu LLM Models}

The Tongyi Qianwen (Qwen) LLM model is  provided by the Alibaba to the open-source community \footnote{https://github.com/QwenLM/Qwen}. It was pre-trained on multilingual data covering various industries and domains, with Qwen-72B being trained over 3 trillion tokens of data.The model can be used for many text understanding and generation tasks, such as  language understanding, coding, reasoning, multilingual capabilities, human preference, agent, retrieval-augmented generation (RAG). Qwen 1.5 is the Beta version of Qwen 2. Qwen 1.5 includes 6 model sizes: 0.5B, 1.8B, 4B, 7B, and 72B, all of which support the context length of 32768 tokens \footnote{https://www.alibabacloud.com/en/solutions/generative-ai/qwen?\_p\_lc=1}. In this paper the Qwen LLM API service is used for inference,
which is free of charge for registered users. 

Ernie LLM is developed by Baidu, which is boosted by a diverse range of training data possessed by Baidu, 
especially that on Chinese language, service applications, and knowledge \footnote{http://research.baidu.com/Blog/index-view?id=183}. 
A feedback and reward mechanism with strategic optimization further enhances the model's capabilities. 
The LLM integrates various types of data and knowledge to automatically generate prompts, which help to generates high-quality results.
It was claimed that the Ernie LLM was used by more tha 200 million users 13 months after launch.
The Baidu LLM model Ernie-4.0 is used in this paper for inference over model web API, which has context length of 8k tokens.
 
 \subsubsection{MiniCPM LLM Model}
MiniCPM-2B is an end-side LLM developed by ModelBest Inc. and TsinghuaNLP, with only 2.4B parameters excluding embeddings \cite{Hu24}.
It has close performance to Mistral-7B on open-sourced general benchmarks with better ability on Chinese, Mathematics and Coding after SFT 
\footnote{https://github.com/OpenBMB/MiniCPM/blob/main/README-en.md}. 
Its overall performance is better than many open source LLM models such as Llama2-13B and Falcon-40B.
The model can be deployed and infer on smartphones.  
It can be finetuend to develop new models with low cost, using a single 1080/2080 GPU for parameter efficient finetuing 
and a 3090/4090 GPU for  full parameter finetuning.
MiniCPM-Llama3-V2.5 is developed based on the Meta open source LLM Llama3 with vision capabilities. 
It achieves state-of-the-art performance on multiple benchmarks among models under 7B parameters. 
In this paper MiniCPM-2B and MiniCPM-Llama3-V2.5 are deployed locally in a desktop computer with Nvidia 1080Ti GPU card.

\section{Experiment Results and Discussions}

\subsection{Experiment Settings}
As the models GPT-3.5 and GPT-4 do not have vision capabilities, only questions from dataset DS-Text are asked to them.
On the other hand, GPT-4o is tested by only questions with images from dataset DS-Image.
Questions without images are not used to test GPT-4o as GPT-4o is expected to have similar performance as GPT-4 with these questions.

In the official UK theory tests, each test has exactly 50 questions. 
While the LLMs can be tested multiple times with 50 questions in each test,
an alternative approach is used in this paper with LLMs being asked all the questions in one test.
The pass decision on the test is then made on the accuracy over all the questions against the passing threshold (86\%).

For all the three tested LLMs, model temperature of 0 is used.
Model temperature determines whether the output is more random or more predictable. 
A lower temperature will result in higher probability, i.e., more predictable outputs.
Other settings for the LLMs include $top\_p=1$, $max\_tokens=100$. 
The prompt sent to the LLMs includes two parts, system prompt and user content.
A specific system prompt setting the role of experience driver to the LLMs is configured as 
``You are an experienced driver to help answer UK driving theory test questions, which have multiple choices but one correct answer.
Your response should include only the first letter (such as A or B) of the correct answer, without any explanation.''
The user content part includes the test question, the test question options and string ``Answer: ''.

\subsection{Test Results and Discussion}

Experiment results for the LLMs on the theory test are presented in Table~\ref{tab:llm-test-results}.

\begin{table*}[ht]
    \centering
    \caption{Comparison of theory test results for the LLMs.}
    \begin{tabular}{llllllll}
    \hline
        Model 	&  \# Questions & \# Correct Answers 	& Accuracy 		& Pass (Y/N) 	& Cost (\$) 	& \# Tokens  	& Time (s) \\ \hline
        GPT-3.5  	&  563  		& 444  				&   79\%  			& No  	&  0.04 		&  75000 		& 360	\\ 
        GPT-4  	&  563  		& 533  				&   95\%  			& Yes  	&  2.05 		&  67590 		& 500	\\ 
        Qwen  	&  563  		& 340  				&   60\%  			& No  	&  0 			&  70002 		& 540	\\        
        Ernie  	&  563  		& 479  				&   85\%  			& No  	&  0 			&  70002 		& 1200	\\  
 MiniCPM-2B  	&  563  		& 319  				&   57\%  			& No  	&  0 			&  70002 		& 210	\\        
MiniCPM-Llama3-V2.5 	&  53  		& 38 				&   72\%  			& No  	&  0	 		&  63102 		& 90		\\ 
        GPT-4o  	&  53  		& 51 				&   96\%  			& No  	&  0.32 		&  63102 		& 180	\\ 
   \hline
    \end{tabular}
    \centering
    \label{tab:llm-test-results}
\end{table*}

In Table~\ref{tab:llm-test-results} columns 2 and 3 shows the number of total test questions and the number of correctly answered questions.
It can be observed that for the test questions without images, the accuracy for GPT-3.5 model is 79\%, 
which is below the passing threshold of 86\% for the UK driving theory test.
As the questions with images are more challenging, GPT-3.5 model may perform worse than on the text only questions.
Compared to GPT-3.5 model, GPT-4 model performs much better with an accuracy of 95\% and passed the theory test.
The result showed improved domain knowledge of GPT-4 model  in driving theory and knowledge as claimed by OpenAI.
But it is also noted that the cost of testing GPT-4 is more than 4 times higher than that for GPT3.5.
Wihile Qwen and MiniCPM-2B models fail the test, with acurracy of 60\% and 57\%, respectively,
Ernie shows a much better performance, with an accuracy of 85\%, which is very close to the passing threshold of 86\%.

For the questions with images, two LLM models GPT-4o and MiniCPM-V2 with vision capabilities are tested.
GPT-4o achieves an impressive accuracy of 96\% over 53 questions, demonstrate great image and scenario understanding capabilities.
On the other hand, MiniCPM-V2 achieves an accuracy of 72\%, which is reasonably good for a small open source LLM.
The advantage of MiniCPM models is the fast response speed and low communication delay when deployed locally.

The above results show that the GPT-4 model can achieve a high accuracy to pass the UK driving theory test (on multiple choice test part),
while the cheaper versions GPT-3.5 and other ones failed the theory test.
The performance of GPT-4o model on questions with images is very impressive.
With respect to communication and computation time,
it takes about 0.7 second, 0.9 second and 3.4 second on average to test GPT-3.5, GPT-4 and GPT-4o for one question, respectively.
The above time covers both communication and LLM computation. 
MiniCPM-2B and MiniCPM-Llama3-V2.5 take less than 0.4 second and 2 seconds for one question respectively.
It can be observed that the time taken on processing one test question will unlikely meet the real-time requirement for driving planning and control,
but it may be acceptable for some driving assistance services such as emergency response. 
 
Furthermore, we check the impact of model temperature and prompt content for OpenAI models.
In the above experiments, model temperature is set to 1 for all three tested LLMs.
For the GPT-3.5 model, an additional configuration of model temperature of 0.7 is also used.
Test results show that a slightly improved accuracy of 80\% (with 450 out of 563 questions correctly answered).
In addition, an additional more general system prompt is tried for GPT-3.5 model.
The new system prompt says  ``You will assist on helping answer UK driving theory test questions, which have multiple choices but one correct answer. 
 Your response should include only the first letter (such as A or B) of the correct answer, without any explanation.''
In this way, the LLM is instructed as a general user instead of experience driver to answer the driving test questions.
With the new system prompt, the accuracy is reduced to 78.3\% (with 441 questions correctly answered), which is not surprising.

\vspace{0.25in}
\section{CONCLUSION}
LLMs hold great potentials for connected autonomous vehicles (CAVs). In this paper we investigated LLMs assisted CAV driving, which can exploit the generalization capabilities of the LLMs. An important question which has not been answered is how safe and qualified is LLM for CAV assistance according to driving theory test.
We tested the three OpenAI LLMs and several other LLMs on understanding  of driving theory knowledgeand skills with multiple choice theory test questions.
Experiment results showed that the fast and less inexpensive models (GPT3.5, Qwen, Ernie and MiniCPM-2B)  failed the theory test. Only model GPT-4 passed the driving theory test with an accuracy of 95\%. 
The GPT-4o and MiniCPM-Llama3-V2.5 models with vision capability achieved an accuracy of 96\% and 72\% over test questions with images. 
GPT-4o demonstrated strong vision capability and potential for autonomous driving assistance. 
The cost and time of using these models to answer the test questions were also measured, 
which showed GPT-4 was much more expensive to use with the benefit of higher accuracy. 
It is noted that that while the GPT-4 model may pass the driving theory test with a passing threshold of 86\%, 
the expectation of the publics on the LLMs accuracy may be much higher for them to provide safety critical driving assistance services. 
Therefore, further performance improvement with the LLMs may still be needed.
In our future work we plan to assess more LLMs on their driving knowledge with the theory test questions. 

\section*{Acknowledgement}

This work was funded by the European Union’s Horizon 2020 research and innovation
programme under the Marie Skłodowska-Curie grant agreement No 824019 and No 101022280,
Horizon Europe MSCA programme under grant agreement number 101086228,
EPSRC with RC Grant reference EP/Y027787/1, and EPSRC/UKRI with grant referenceRCP 15831/DCM4480.
The authors also thank Driving Instructor Websites for sharing their driving theory test questions.

\vfill
\end{document}